\documentclass[11pt,a4paper]{article}
\usepackage[hyperref]{emnlp2020}
\usepackage{times}
\usepackage{latexsym}

\usepackage{graphicx}
\usepackage{enumerate}
\usepackage{booktabs}
\usepackage{tabularx}
\usepackage{multirow}
\usepackage{pdflscape}
\usepackage{afterpage}
\usepackage{rotating}
\usepackage{xcolor}
\usepackage{tcolorbox}
\usepackage{array}
\usepackage[colorinlistoftodos]{todonotes}
\newcolumntype{H}{>{\setbox0=\hbox\bgroup}c<{\egroup}@{}}
\definecolor{mygreen}{HTML}{3cb44b}

\usepackage{microtype}

\aclfinalcopy 

\title{Template Guided Text Generation for Task-Oriented Dialogue}

\author{Mihir Kale \\
  Google  \\
  Mountain View \\
  \texttt{mihirkale@google.com} \\\And
  Abhinav Rastogi \\
  Google Research \\
  Mountain View \\
  \texttt{abhirast@google.com} \\}

\date{}

\begin{document}
\maketitle
\begin{abstract}
Virtual assistants such as Google Assistant, Amazon Alexa, and Apple Siri enable users to interact with a large number of services and APIs on the web using natural language. In this work, we investigate two methods for Natural Language Generation (NLG) using a single domain-independent model across a large number of APIs. First, we propose a schema-guided approach which conditions the generation on a schema describing the API in natural language. Our second method investigates the use of a small number of templates, growing linearly in number of slots, to convey the semantics of the API. To generate utterances for an arbitrary slot combination, a few simple templates are first concatenated to give a semantically correct, but possibly incoherent and ungrammatical utterance. A pre-trained language model is subsequently employed to rewrite it into coherent, natural sounding text. Through automatic metrics and human evaluation, we show that our method improves over strong baselines, is robust to out-of-domain inputs and shows improved sample efficiency. \footnote{Our code and data is available at \href{https://github.com/google-research/schema-guided-dialogue}{github.com/google-research/schema-guided-dialogue}}
\end{abstract}

\section{Introduction}
Virtual assistants have become popular in recent years and task-completion is one of their most important aspects. These assistants help users in accomplishing tasks such as finding restaurants, buying sports tickets, finding the weather etc., by providing a natural language interface to many services or APIs available on the web. Most systems include a natural language understanding and dialogue state tracking module for semantic parsing of the dialogue history. This is followed by a policy module which interacts with the APIs, whenever required, and generates the actions to be taken by the system to continue the dialog. In the end, the Natural Language Generation (NLG) module converts these actions into an utterance, which is surfaced to the user. Being the user-facing interface of the dialogue system, NLG is one of the most important components impacting user experience.



\begin{figure*}[ht]
\centering
\includegraphics[width=0.95\textwidth]{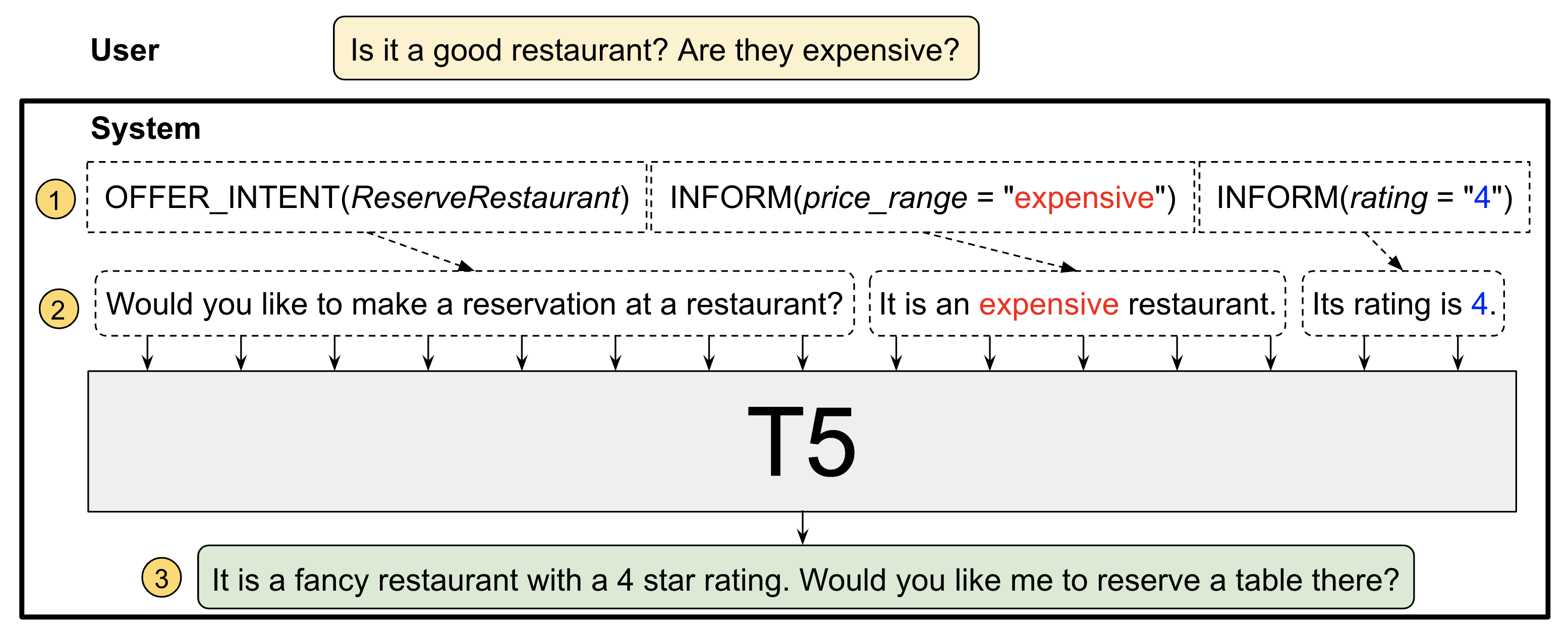}
\caption{Overall architecture of our proposed template guided approach. 1. The policy module outputs a set of actions in response to the user utterance. 2. Simple templates convert each action into a natural language utterance. 3. Template-generated utterances are concatenated and fed to a T5 encoder-decoder model\cite{raffel2020exploring}. The model rewrites it to a conversational response surfaced to the user.}
\label{fig:overall}
\end{figure*}

Traditional NLG systems heavily utilize a set of templates to produce system utterances. Although the use of templates gives good control over the outputs generated by the system, defining templates becomes increasingly tedious as more APIs are added. Supporting multi-domain conversations spanning multiple APIs quickly grows out of hand, requiring expert linguists and rigorous testing to ensure the grammatical correctness and appropriateness of generated utterances. Consequently, data-driven generative approaches have gained prominence. Such systems require much less effort and can generate utterances containing novel patterns. Meanwhile, with the rapid proliferation of personal assistants, supporting large number of APIs across multiple domains has become increasingly important, resulting in research on supporting new APIs with few labelled examples (few-shot learning). To this end, generative models pre-trained on large amounts of unannotated text have been increasingly successful. 

In this work, we address the challenges of joint modeling across a large number of domains, and data efficient generalization to new domains and APIs for NLG. Our contributions are the following:
\begin{enumerate}
    \item We propose two methods for zero-shot and few-shot NLG. Our first method, the Schema-Guided NLG, represents slots using their natural language descriptions. Our second method - Template Guided Text Generation (T2G2) employs a simple template-based representation of system actions and  formulates NLG as an utterance rewriting task (Figure \ref{fig:overall}). 
    \item We present the first NLG results on the Schema-Guided dialogue dataset \cite{rastogi2019scalable},  which  exceeds all other datasets in scale, providing a total of 45 APIs over 20 domains. While the current state-of-the-art pre-training based methods struggle to generalize to unseen (zero-shot) APIs, our proposed methods are robust to out-of-domain inputs and display improved sample efficiency.
    \item We conduct an extensive set of experiments to investigate the role of dialogue history context, cross-domain transfer learning and few-shot learning. We share our findings to guide the design choices in future research.
\end{enumerate}

\section{Related Work}
Natural language generation from structured input (NLG) has been an active area of research, facilitated by creation of datasets like WikiBio \cite{lebret2016neural}, E2E challenge \cite{novikova2017e2e}, WebNLG \cite{gardent2017webnlg} and MultiWOZ \cite{budzianowski2018multiwoz}. Neural sequence models have been extensively used in a variety of configurations for NLG in dialogue systems.  \citet{wen2017network} proposed a two-step approach: first generating a delexicalized utterance with placeholders for slots and then post-processing it to replace placeholders with values from API results, whereas \citet{nayak2017plan} highlighted the importance of conditioning responses on slot values. 

Sequence to sequence architectures directly converting a sequential representation of system actions to a system response are also very common \cite{wen2015semantically,duvsek2016sequence,zhu2019multi,chen2019semantically}. Domain-adaptation and transfer learning in low resource settings has also been an extensively studied problem \cite{tran2018adversarial,chen-etal-2020-shot, peng2020few, mi2019meta}, with recently released datasets like SGD \cite{rastogi2019scalable} and FewShotWOZ \cite{peng2020few} providing a good benchmark. Meanwhile, language models pre-trained on large amount of unannotated text corpus have achieved state-of-the-art performance across several natural language processing tasks \cite{devlin2019bert, yang2019xlnet, liu2019roberta, radford2019language, keskar2019ctrl}, including natural language generation \cite{peng2020few,kale2020machine}.

Our template based approach bears similarities to sentence fusion \cite{barzilay2005sentence}, and prototype based text editing \cite{hossain2020simple,cao2018retrieve,guu2018generating,wu2019response}. However, none of these works tackle text generation from structured data.

\section{Model}
\label{sec:model}

\begin{figure*}[t]
\centering
\begin{tabular}[t]{ c | c }
    \textbf{Approach} & \textbf{Representation of System Actions} \\\hline
    Naive & inform ( restaurant = Opa! ) inform ( cuisine = greek ) \\\hline
    Schema Guided & inform ( name of restaurant = Opa! ) inform ( type of food served = greek ) \\\hline
    Template Guided & How about the restaurant Opa!. The restaurant serves greek food.\\\hline
    Ground Truth & Opa! is a nice greek restaurant. How does it sound?
\end{tabular}
    \caption{An example showing the representation of system actions utilized by the three schemes. The template representation is generated by concatenating sentences obtained from two templates, which are ``\textit{inform(restaurant = \$x) $\rightarrow$ How about the restaurant \$x.}" and ``\textit{inform(cuisine = \$x) $\rightarrow$ The restaurant serves \$x food.}".}
    \label{fig:encodings}
\end{figure*}

For a given system dialogue turn, let $\mathcal{A} = \{ d_i(s_i=v_i) \}_{i=1}^A$ be the set of actions which are produced by the system, where $A$ is the total number of actions for this turn. Each action consists of a single dialogue act $d_i$ representing the semantics of the action, along with optional slot and value parameters - $s_i$ and $v_i$ respectively. For example, \textit{inform}, \textit{req\_more} and \textit{request} are some of the dialogue acts defined in the SGD dataset \cite{rastogi2019scalable}, which are used for informing the value of a slot to the user, asking if the user needs some other help, and requesting the value of a slot from the user respectively. Some acts like \textit{inform} require both the slot and value parameters, whereas acts like \textit{request} require the slot parameter only and acts like \textit{req\_more} require none. Some datasets allow multiple slot-value arguments for a single act, but such actions can generally be converted to the above representation by decomposing them into multiple actions with the same act, each containing exactly one slot-value pair.

The goal of NLG is to translate $\mathcal{A}$ to a natural language response with the same semantic content. To this end, we first convert the set $\mathcal{A}$ into a sequence. Then, we finetune a  Text-to-Text Transfer Transformer (T5) \cite{raffel2020exploring} model, which is a pre-trained sequence to sequence transformer, to generate the natural language response using this sequence as input. Now, we present three different methods for converting $\mathcal{A}$ into a sequence, the last two being our contributions. They are also summarized in Figure \ref{fig:encodings}.

\subsection{Naive Representation}
This approach uses the most basic representation of actions, similar to that used in many prior works \citep{novikova2017e2e, zhu2019multi, peng2020few}. Canonical representations of each action - $a_i, a_i(s_i)$ or $a_i(s_i=v_i)$, depending on the parameters present in the action, are concatenated together to obtain a sequence representation of $\mathcal{A}$. Although this representation is simple to obtain and gives state of the art results for several data-to-text benchmarks \cite{kale2020text}, it suffers from two drawbacks -
\begin{enumerate}[(i)]
    \item \textbf{Semantics - } This representation doesn't convey much information about the semantics of a slot. Consequently, the model may need a larger number of training examples to identify the semantics of a slot from its usage in the system utterances in the training data.
    \item \textbf{Representation Bias -} This representation is very different from what the encoder has seen during pre-training phase, which is natural language text. As a result, the representations learnt during pre-training may not transfer well. \citet{peng2020few} mitigate this by conducting additional pre-training using large scale annotated dialogue datasets. While this method is effective, a large in-domain corpus may not always be available.
\end{enumerate}

\subsection{Schema Guided Representation}
Recent work on low-resource natural language understanding tasks have used natural language descriptions of slots. These descriptions are easy to obtain, directly encode the semantics of the slot and have been shown to help when in-domain training data is sparse. While description based representations have become popular for tasks like spoken language understanding \citep{bapna2017towards} and dialogue state tracking \citep{rastogi2019scalable}, they have not yet been applied to the language generation task. We propose an extension of the Naive representation by replacing the slot names with their natural language descriptions. The action representations, as illustrated in Figure \ref{fig:encodings}, are $a_i, a_i(desc(s_i))$ and $a_i(desc(s_i)=v_i)$, where $desc(s)$ represents a natural language description of slot $s$. This solves the first drawback of the Naive representation mentioned above.

\subsection{Template Guided Representation}
\label{sec:t2g2}

\begin{figure}[t]
\centering
\begin{tabular}[t]{ p{2.7cm} | p{4cm} }
    \textbf{Action} & \textbf{Template} \\\hline
    \textit{notify\_success} & Your ride is booked and the cab is on its way. \\
    \textit{goodbye} & Have a safe ride!\\ \hline
    \textit{request(dest)} & Where are you riding to?\\
    \textit{request(shared)} & Are you comfortable sharing the ride?\\ \hline
    \textit{confirm(dest=}\$x\textit{)} & You are going to \$x. \\
    \textit{inform(fare=}\$x\textit{)} & Your ride costs \$x dollars. \\
    \textit{inform(seats=}\$x\textit{)} & The cab is for \$x riders.
    \end{tabular}
    \caption{Example templates for a ride-sharing API. Parameterized templates are defined for actions which contain a slot value.}
    \label{fig:templates}
\end{figure}

We solve the representation bias problem by converting the set of actions output by the system into a natural language utterance. We employ a technique similar to that used in \citet{rastogi2019scalable}, where simple utterances are generated using a minimal set of manually defined templates. Specifically, as shown in Figure \ref{fig:templates}, we define one template for each system action. The representation of $\mathcal{A}$ is obtained by concatenating the corresponding templatized representation of each action in $\mathcal{A}$. See Figure \ref{fig:encodings} for a complete example.

Note that, our focus here is not to generate conversational and grammatically correct utterances, but to have a simple representation of the actions, which can be rewritten by the model into a natural and fluent response. Hence, we do not need to cover all edge cases typically required in template based methods - handling of plurals, subject-verb agreement, morphological inflection etc. - and only need to define a small number of templates. For most APIs, this amounts to around 15-30 templates, which can easily be written by the API developer. The actual number varies depending on the number of slots and intents supported by the API \footnote{Please see Appendix \ref{template-examples} for more examples of templates.}. Some special slots like date, time and price are formatted using special rules, which can be reused across APIs. For instance, we convert the date ``2019-03-06" to ``6th March", the time ``18:40" to ``6:40 pm", and price ``60" to ``\$60". We call this step \textit{value paraphrasing}. Since this method relies on a combination of templates and transfer learning from language models, we name it \textbf{Template Guided Text Generation (T2G2)}.

\section{Experimental Setup}
\label{experimental-setup}
We conduct a series of experiments to compare the three system action representations presented above. We also evaluate NLG in few-shot settings and investigate a few other aspects of the SGD dataset. In each of the experiments reported in this paper, we start with a pre-trained T5-small model\footnote{\href{https://github.com/google-research/text-to-text-transfer-transformer}{github.com/google-research/text-to-text-transfer-transformer}}. It has 6 layers each in the encoder and decoder, with a total of around 60 million parameters. The model is then fine-tuned on the corresponding dataset using a constant learning rate of 0.001 and batch size of 256 for 5000 steps. The checkpoint yielding the highest BLEU score on the development set is picked for reporting test set results. During inference, we use beam search with a width of 4 and length penalty $\alpha = 0.6$.

\section{Action Representations}
\label{sec:representations}

\begin{figure}[t]
\centering
\includegraphics[width=0.48\textwidth]{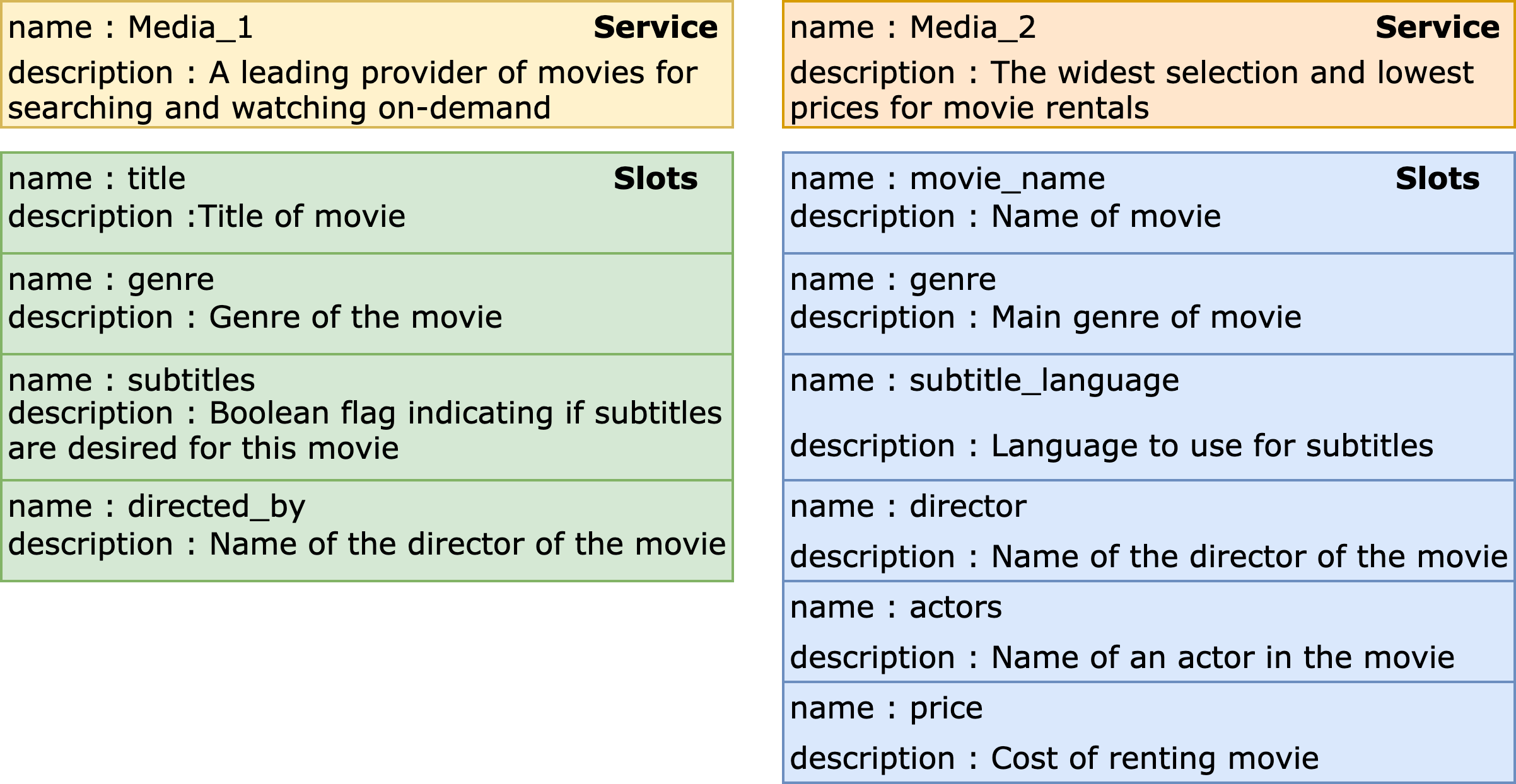}
\caption{Schemas of two APIs from the Media domain present in the SGD dataset.}
\label{fig:schema_example}
\end{figure}

We compare the different methods of action representation on MultiWOZ 2.1 \cite{budzianowski2018multiwoz}, the cleaned version of the E2E restaurant corpus \cite{novikova2017e2e, duvsek2019semantic} and the Schema-Guided Dialogue (SGD)~\cite{rastogi2019scalable} dataset. The SGD dataset features a larger number of domains and slots, and the presence of multiple APIs per domain (Figure~\ref{fig:schema_example}) makes it representative of practical scale-related challenges faced by today's virtual assistants. Furthermore, as opposed to the other two datasets, its evaluation sets contain many domains, and consequently slots, which are not present in the training set. Even for domains shared between the training and evaluation sets, the evaluation sets contain additional slots in some cases. This focus on zero-shot generalization to new domains and APIs makes SGD more challenging than existing NLG benchmarks. Table \ref{datastats} compares these datasets.

\begin{table}[]
\centering
\begin{tabular}{lccc} \hline
\textbf{Statistic}      & \textbf{E2E}  & \textbf{MWoz} & \textbf{SGD}  \\ \hline
Domains        & 1    & 7        & 20   \\
Unseen domains & 0    & 0        & 4    \\
System acts    & 1    & 7        & 10   \\
Slots          & 8    & 23       & 184  \\
Unseen slots   & 0    & 0        & 41   \\
Train size     & 33k  & 57k      & 160k \\
Dev size       & 4.3k & 7.3k     & 24k  \\
Test size      & 4.7k & 7.3k     & 42k \\ \hline
\end{tabular}
\caption{Comparison of NLG datasets. MWoz is short for MultiWOZ. Train/Dev/Test sizes represent the number of system turns. Unseen domains refers the test set.}
\label{datastats}
\end{table}

\begin{figure*}[t]
\centering
\begin{minipage}{15.6cm}\vspace{0mm}    \centering
\begin{tcolorbox}[colback=yellow!5!white]
    \centering
    \scriptsize
    \begin{tabular}{lp{0.85\columnwidth}}
    \textbf{Model} & \textbf{Input or Generated sequence} \\
    \midrule
    \textit{Actions} & Flights offer(airlines=American Airlines, outbound\_departure\_time=2:40 pm, is\_nonstop=True, price=\$78) \\[0.5pt]
    \textit{Template} & Would you like to fly with American Airlines? The onward flight takes off at 2:40 pm. It is a direct flight. The ticket costs \$78. \\[0.5pt]
    \textit{Reference} & Do you wish to fly on American Airlines? The onward direct flight leaves at 2:40 pm, and the ticket price is \$78. The ticket costs \$78 \\[2.5pt]
    \textbf{Naive} & There is an American Airlines flight that leaves at 2:40 pm and is \underline{not stop}. The ticket costs \$78.
    \\[0.5pt]
    \textbf{Schema} & There is an American Airlines flight that leaves at 2:40 pm and is \underline{not stop}. The total cost is \$78.
    \\[0.5pt]
    \textbf{T2G2} & How about an American Airlines flight that leaves at 2:40 pm? It's a direct flight and costs \$78. \\[2.5pt]
    \midrule
    \textit{Actions} & RentalCars confirm(has\_insurance=True, car\_type=hatchback) \\[0.5pt]
    \textit{Template} & Please confirm the following details: It is a Hatchback car. It comes with insurance.  \\[0.5pt]
    \textit{Reference} & Please confirm: You want me to book a Hatchback car with insurance. \\[2.5pt]
    \textbf{Naive} & Please confirm: You want me to reserve a Hatchback car with a \underline{reassurance}. \\[0.5pt]
    \textbf{Schema} & Please confirm: You want to rent a Hatchback. \\[0.5pt]
    \textbf{T2G2} & Okay, please confirm that you'd like to have a Hatchback car with an insurance. \\[2.5pt]
    \midrule
    \textit{Actions} & Media inform(starring=Andrea Martin) \\[0.5pt]
    \textit{Template} & Antonio Bustorff acted in it. \\[0.5pt]
    \textit{Reference} & The actor is Antonio Bustorff. \\[2.5pt]
    \textbf{Naive} & The movie is \underline{directed} by Antonio Bustorff. \\[0.5pt]
    \textbf{Schema} & The movie is \underline{directed} by Antonio Bustorff. \\[0.5pt]
    \textbf{T2G2} & Antonio Bustorff acted in the movie. \\[2.5pt]
 \vspace{-5mm}
    \end{tabular}
\end{tcolorbox}
\vspace{-2mm}
\caption{A few examples of utterances generated from different models, along with the system actions, their template based representation (as shown in Figure~\ref{fig:overall}) and the reference output. The errors are underlined.}
\label{examples}
\end{minipage}
\end{figure*}

\subsection{Automatic Evaluation}

Following prior work \cite{wen2015semantically}, we use BLEU \citep{papineni2002bleu} and Slot Error Rate (SER) \citep{duvsekneural} as automatic metrics. SER represents the fraction of generated texts where at least one slot was not correctly copied from the structured data. Since this metric relies on string matching, we cannot use it to evaluate binary slots like \textit{has\_live\_music}. Its exact match nature also prevents it from identifying paraphrases of slot values, e.g. \textit{expensive} and \textit{costly}. 
For E2E we use additional metrics used in prior work for this benchmark - NIST \citep{doddington2002automatic}, ROUGE-L \citep{lin2004rouge}, METEOR \citep{lavie2007meteor}, CIDEr \citep{vedantam2015cider}, and BLEU.

\begin{table}[]
\centering
\begin{tabular}{lcc} \hline
\textbf{Model}  & \textbf{BLEU}  & \textbf{SER}   \\ \hline
\textbf{HDSA} \cite{chen2019semantically}     & 26.5 & 12.14  \\
\textbf{SC-GPT} \cite{peng2020few}   & 30.8 & \textbf{0.53} \\
\textbf{Naive}   & \textbf{34.6}  & 1.27    \\
\textbf{Schema}   & 33.3  & 1.89    \\
\textbf{T2G2} & 34.4 & 1.85    \\ \hline
\end{tabular}
\caption{Performance of models on MultiWOZ.}
\label{results-multiwoz}
\end{table}

\begin{table}[]
\centering
\begin{tabular}{lccccc} \hline
\textbf{Model}   & BLEU  & N & M & R & C \\ \hline
\textbf{SC-LSTM}  & 23.7 & 4.0 & 32.9  & 39.3   & 0.4  \\
\textbf{TGen}     & 40.7 & 6.2 & 37.8  & 56.1   & \textbf{1.9}  \\
\textbf{Naive}   & 42.1 & \textbf{6.4} & 38.5  & 56.2   & \textbf{1.9}  \\
\textbf{Schema}   & \textbf{43.1} & \textbf{6.4} & \textbf{38.7}  & 56.8   & \textbf{1.9}  \\
\textbf{T2G2}    & 42.5 & \textbf{6.4} & \textbf{38.7}  & \textbf{56.9}   & \textbf{1.9} \\ \hline
\end{tabular}
\caption{Performance of models on E2E. Results for SC-LSTM \cite{wen2015semantically} and TGen \cite{novikova2017e2e} have been taken from \citet{duvsek2019semantic}. N,M,R,C stand for NIST, METEOR, ROUGE and CIDEr respectively.} 
\label{results-e2e}
\end{table}

\paragraph{MultiWOZ and E2E} Table \ref{results-multiwoz} lists results on the MultiWOZ and Table \ref{results-e2e} on E2E. We train separate models for each dataset. On both datasets, T2G2 and Schema are comparable to the state-of-the-art Naive approach. We note that the SER score on MultiWOZ is slightly worse in comparison with SC-GPT.  SC-GPT generates 5 predictions for each input and then ranks them based on the SER score itself. On the other hand, we generate a single output, on which SER is evaluated. Overall, the results indicate that with enough annotated data, the Naive approach is enough to attain good performance. Both datasets are large and feature limited variety (MultiWOZ has 57K utterances spread over just 5 domains, while E2E has 33k utterances spread over just 8 slots). Zero-shot and few-shot settings offer a greater and more realistic challenge, and we explore these settings next. The SGD dataset, which spans 20 domains, enables us to study these settings.

\begin{table}[h]
\centering
\begin{tabular}{lc|c|c|c}
\hline
\textbf{BLEU} & \textbf{Naive}  & \textbf{Schema} & \textbf{T2G2} & \textbf{Copy} \\ \hline
Unseen  & 14.9 & 15.8 & \textbf{22.2} & 16.1  \\
Seen    & 27.7 & 27.5 & \textbf{29.4} & 19.2  \\
Overall & 26.2 & 26.2 & \textbf{28.6} & 18.8  \\ \hline \hline
\textbf{SER} & \textbf{Naive}  & \textbf{Schema} & \textbf{T2G2} & \textbf{Copy} \\ \hline
Unseen  &  0.7 &  0.4 &  \textbf{0.0} &  -  \\
Seen    &  1.1 &  0.8 &  \textbf{0.4} &  -  \\
Overall &  1.0 &  0.8 &  \textbf{0.4} &  -  \\ \hline
\end{tabular}
\caption{BLEU and SER metrics on SGD dataset. \textit{Copy} refers to a trivial baseline comprising of the template based input representation and has 0 SER by definition.}
\label{results-sgd}
\end{table}

\paragraph{Adaptation to New Domains} The ideal NLG model should be able to handle domains it was not exposed to during training. 
The SGD dataset, which features unseen domains in the evaluation sets, lets us us assess the zero-shot capability of NLG systems. We report results in Table \ref{results-sgd} on two test sets - the \textit{seen} set consists of domains that were seen during training, while the \textit{unseen} set consists of brand new domains aka the zero-shot setting. Firstly, all models exhibit low SER scores in both seen and unseen domains, with the template approach being the lowest. This suggests that pre-trained language models are adept at copying and this skill also generalizes to out-of-domain examples. 

The Schema-Guided representation performs at par with Naive representation on seen domains. At the same time, the slot descriptions do improve performance on the unseen domains (+0.9 BLEU), albeit to a limited degree. More effective ways of incorporating descriptions is a promising area for future work. For the seen domains, T2G2 outperforms Naive by 1.7 BLEU. The results on the unseen domains are more striking with an improvement of 7.3 points. This confirms the hypothesis that our simple template based input scheme offers superior generalization capabilities with a low overhead. The template model learns to "fuse" sentences and is able to successfully extend this skill to unseen domains. 

\subsection{Qualitative Analysis}
In Figure~\ref{examples} we list a few examples of model predictions. The first example illustrates a case where the model has to deal with a seen domain \textit{Flights} but an unseen slot \textit{is\_nonstop}. Such a case would be common when new functionality needs to be added to an existing domain. Both Naive and Schema are unable to verbalize the slot correctly. While the template input contains all the information, it sounds very robotic. T2G2, on the other hand, takes the 4 template sentences as input and rewrites them into a fully accurate but much more natural sounding response.

The next example is from \textit{RentalCars}, and features an unseen slot \textit{has\_insurance}. Schema fails to mention this slot. Naive attempts to verbalize it, but uses the wrong word (\textit{reassurance}). T2G2, however, is able to paraphrase the template input into grammatical text without dropping any information.

The final example features an unseen slot \textit{starring} from the \textit{Movies} domain. Naive and Schema treat Antonio Bustroff as a director, since the slot \textit{directed\_by} appears during training. However, T2G2 simply relies on the template input and copies the phrase \textit{acted in}. We refer the reader to Appendix \ref{qualitative-examples} for more qualitative examples.

\subsection{Human Evaluation} 

We conduct a human evaluation study via crowd sourcing \footnote{Examples of the rating UI can be found in Appendix \ref{human-eval-tasks}.}. Each human rater is shown the responses generated by different models and the ground truth response in a random order. Following \citep{peng2020few}, they are asked to rate each response on a scale of 1 (bad) to 3 (good) along two axes - \textit{informativeness} and \textit{naturalness}. Informativeness quantifies whether the response contains all the information contained in the dialogue acts, whereas naturalness evaluates whether the response sounds coherent, grammatical and natural. Each example is rated by 3 different workers. The final metric is an average of all the ratings. 

\begin{table}[h]
\begin{tabular}{ll|l|l|l}
\hline
\multicolumn{5}{c}{\textbf{Naturalness}} \\\hline
& Naive & Schema & T2G2 & GT \\\hline
Unseen & 2.43$^{4}$ &	2.41	& \textbf{2.46}$^{2,4}$	& 2.37 \\
Seen & \textbf{2.48}$^{4}$	& 2.45 &	2.47$^{4}$	& 2.40 \\
Overall & 2.45$^{4}$	& 2.43$^{4}$ &	\textbf{2.46}$^{2,4}$ &	2.38 \\ \hline \hline
\multicolumn{5}{c}{\textbf{Informativeness}} \\\hline
& Naive & Schema & T2G2 & GT \\\hline
Unseen & 2.36	& 2.49$^{1}$	& \textbf{2.55}$^{1,2}$	& 2.51$^{1}$ \\
Seen & 2.57	& \textbf{2.59}$^{4}$	& 2.56	& 2.54 \\
Overall & 2.46	& 2.54$^{1}$	& \textbf{2.56}$^{1}$	& 2.53$^{1}$ \\ \hline
\end{tabular}
\caption{Human evaluation results comparing different models and the ground truth. The superscripts 1 to 4 indicate that the model is significantly better than Naive, Schema, T2G2 and ground truth respectively, as determined by a one-tailed paired t-test with $p < 0.05$.} 
\label{results-human}
\end{table}

A total of 500 randomly chosen examples are rated - 250 each from seen and unseen domains - across the 3 models discussed above and the ground truth response (\textit{human}). With 3 ratings per example, this leads to a total of 6,000 ratings. 
Results are shown in Table \ref{results-human}. \\
\textbf{Naturalness} On the overall test set, all models outperform the human authored ground truth. This showcases the strength of pre-trained language models in generating natural sounding utterances, echoing findings from prior works.  \cite{radford2019language, peng2020few}. \\
\textbf{Informativeness} Simply generating a fluent response is not enough. Its paramount for the responses to be factually grounded in the structured data, so that the wrong information is not conveyed to the user. For informativeness, we notice that all models perform well on the seen domains. However, on unseen domains,  the Naive approach fares poorly. Schema outperforms Naive by a large margin on unseen domains. T2G2 further improves upon Schema. These results suggest Schema and T2G2 offer promising avenues to improve the zero-shot generalization capability of NLG systems. 
Moreover, both Naive and Schema see large drops on unseen domains, while T2G2 performs equally well on both seen and unseen domains.

Recall that Naive representation demonstrated strong scores on the SER metric for unseen domains. However, the low human scores on informativeness suggest that getting perfect scores on metrics like SER may not be a reliable way to judge factual accuracy. As models become stronger, better evaluation metrics need to be developed to accurately measure the improvements. 

\section{Few-Shot NLG}
Virtual assistants need to support a constantly increasing number of domains and APIs. In order to keep labelled data costs under control, improving few-shot learning methods is important. In this section, we study the trade-off between the number of annotated training examples and performance of NLG.

\subsection{Dataset}

\begin{table}[h] 
\centering
\begin{tabular}{lcc} \hline
$K$   & Dialogues & Examples \\ \hline
5   & 70        & 558      \\
10  & 140       & 1,075     \\
20  & 280       & 2,140     \\
40  & 560       & 4,312     \\
80  & 1,120      & 8,624     \\
All & 16,141     & 164,978  \\ \hline
\end{tabular}
\caption{Data statistics of FewShotSGD training splits.}
\label{fewshotsgd-stats}
\end{table}

Prior work \cite{mi2019meta, tran2018adversarial, wen2016multi} has studied few-shot learning and domain adaptation in a simulated setting by creating small subsets. However, lack of knowledge of the exact data splits makes it difficult to make comparisons to other methods. To remedy this, we create a new canonical split of the SGD dataset as described below. 
\begin{itemize}
\item We make $K$-shot subsets for varying values of $K$ $[5,10,20,40,80]$. In this setting each of the 14 domains from the training set have $K$ dialogs.
\item For all the few-shot splits we make sure that they contain examples for every dialogue act and slot type present in the full training set. For every domain, we make sure that each dialog act (inform, request etc.) and slot (name, time, price etc.) is represented at least once. However, all combinations of dialog acts and slots may not exist.
\item The dev and test sets are left untouched. 
\end{itemize}
This benchmark is referred to as FewShotSGD and we make the exact splits publicly available. The exact number of examples in each split is given in Table \ref{fewshotsgd-stats}.

\subsection{Results}

\begin{figure}[t]
\includegraphics[width=0.48\textwidth]{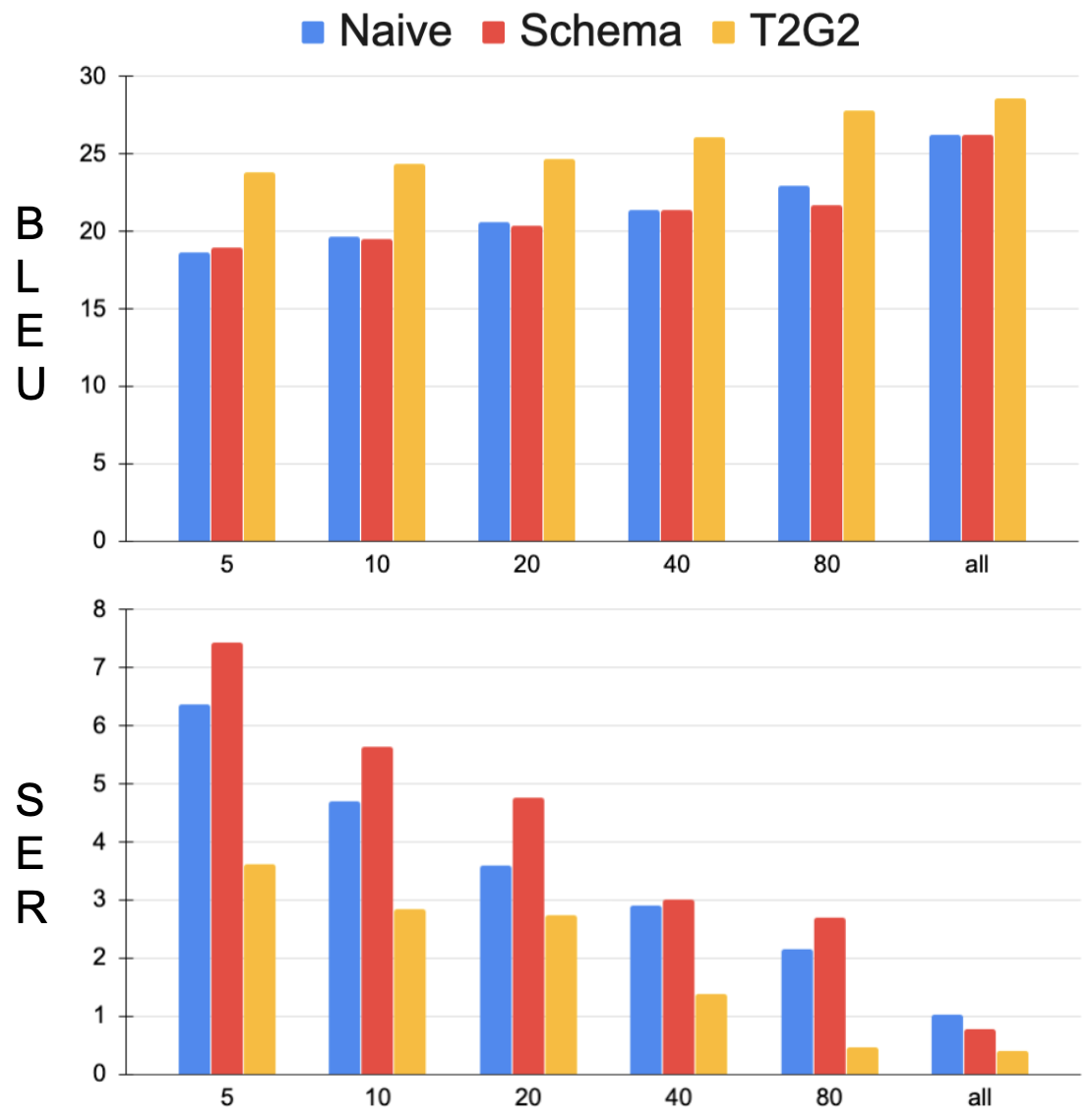}
\caption{Performance in few-shot settings. The x-axis indicates the number of dialogues per domain in the training set. For exact scores, please refer to Appendix \ref{exp-details}.}
\label{results-fewshot}
\end{figure}

In few shot experiments, we examine the performance of different models as a function of the amount of labelled data. The training setup remains the same, as described in section \ref{experimental-setup}. Results are reported in Figure~\ref{results-fewshot}, where we can clearly see the performance improving as more training data becomes available. In all the $K$-shot settings,  T2G2 gives consistent improvements of 4-5 BLEU while reducing the SER by a large margin. Even in the extreme 5-shot setting, the SER is just 3.6\%. Remarkably, T2G2 in the 80-shot setting outperforms the Naive model trained on the \textsl{entire} dataset, which is ~20x larger. In the 5-shot setting, T2G2 performs on par with 80-shot Naive. We take this as evidence that our template guided input representation can lead to significant reduction in labelled data requirements.

\section{Other Experiments}
In this section, we conduct experiments to explore a few other aspects of our setup on the SGD dataset. For these experiments we use the Naive representation, since it is more widely adopted in prior work. We hope that these experiments will guide design choices in the future NLG models.


\subsection{Joint Modeling}
\label{joint-modeling}
Joint modeling, instead of domain specific models, could be beneficial in low resource settings if there is some similarity between the underlying structure. Furthermore, having a single model for all domains also reduces the maintenance workload and is resource efficient. For NLG systems, it could also help in maintaining consistent styles across domains and APIs. 

\begin{table}[]
\centering
\begin{tabular}{l|ll|ll} \hline
\textbf{Domain} & \multicolumn{2}{c|}{\textbf{Separate}} & \multicolumn{2}{c}{\textbf{Joint}}  \\ 
& \textbf{BLEU} & \textbf{SER} & \textbf{BLEU} & \textbf{SER} \\ \hline
Homes       & 22.9        & 1.6        & 26.3       & 0.2         \\
Buses       & 18.6        & 4.0        & 23.4       & 0.0         \\
Media       & 28.9        & 8.4        & 29.9       & 4.6            \\
RideShare   & 20.3        & 2.1        & 26.0       & 0.0         \\
Movies      & 21.4        & 23.0       & 29.3       & 4.9         \\
Flights     & 19.7        & 1.0        & 20.5       & 0.0            \\
Music       & 25.3        & 0.6        & 28.5       & 0.0            \\
Services    & 25.3        & 0.6        & 29.2       & 0.0         \\
RentalCars  & 17.6        & 9.2        & 22.1       & 2.0            \\
Restaurants & 25.8        & 4.0        & 27.5       & 0.1         \\
Events      & 30.7        & 0.5        & 31.9       & 0.0         \\
Hotels      & 26.6        & 1.6        & 29.7       & 0.2         \\ \hline
\textbf{Average} & 23.6  & 4.7 & \textbf{27.0} & \textbf{1.0} \\ \hline
\end{tabular}
\caption{Joint vs domain-specific (separate) NLG.} 
\label{results-joint}
\end{table}

Because of these merits, we investigate the effect of joint modeling on SGD dataset. We focus on the 12 domains that are present in all 3 splits - train, dev and test. We train a single model on all these domains and compare it with individual models trained for each domain separately. As shown in Table~\ref{results-joint}, joint modeling leads to a win-win situation by improving BLEU by 3.4 points and reducing SER from 4.7\% to just 1\%, while requiring fewer parameters and resources. For further analysis of transfer learning across domains, we refer the reader to Appendix \ref{transfer-learning}.

\subsection{Role of Context}
\begin{table}[t]
\centering
\begin{tabular}{cccccc} \hline
\multicolumn{1}{l}{$k$} & \multicolumn{1}{l}{0} & \multicolumn{1}{l}{1} & \multicolumn{1}{l}{3} &  \multicolumn{1}{l}{5} & \multicolumn{1}{l}{7} \\ \hline
BLEU & 26.2 & 29.0 & 31.5 & 32.4 & 32.6\\ 
SER  & 1.0  &  1.0 &  0.8 &  0.9 &  0.7\\ \hline
\end{tabular}
\caption{Changing the size of the context. \textit{k} represents the number of previous utterances used.}
\label{table:context}
\end{table}
Dialogue acts represent the semantic content of the system response, but they don't contain any information about the lexical and syntactic content. 
The previous utterances in the dialogue history or context are important for generating good responses because they can help model conversational phenomena such as co-reference, elision, entrainment (lexical and syntactic alignment of responses) and avoid repetition \cite{duvsek2016context}.  Context also helps add variations to the responses generated across different conversations for the same system actions.

Table \ref{table:context} shows the performance of NLG as more utterances from the dialogue context are given as input. In these experiments, we concatenate the last $k$ utterances to the system action representation obtained from the Naive method. The model benefits from the additional context, showing an improvement of upto 6 BLEU. Just a single context utterance - the previous user utterance - results in an improvement of nearly 3 BLEU.

The evaluation for $k>=2$ is not completely realistic, because we used the ground truth system utterances in the context during evaluation as opposed to the utterances generated by the NLG model itself. Regardless, the improvements clearly point to effectiveness of the added context at the cost of more resources. We hope these results inspire more work in this exciting direction.

\section{Conclusion and Future Work}
In this work, we proposed schema guided and template guided input representation schemes for task oriented response generation. Coupled with pre-trained language models, the template guided approach enables zero-shot generalization to new domains with little effort. Moreover, we show that it can lead to drastic reduction in annotation costs. We also present the first set of results on the multi-domain SGD dataset, which we hope will pave the way for further research in few-shot, zero-shot and multi-domain language generation.

While in this paper we use standard pre-trained models, designing pre-training tasks tailored to \textit{sentence fusion} is an interesting line of future work. We also hope to apply T2G2 to languages other than English. Obtaining annotated data in non-English languages is an even bigger challenge, making the sample efficiency of our template rewriting approach especially suited to this setting. Another interesting line of future work is to investigate the use of T2G2 for generating user utterances, which could be useful for dialogue data augmentation and user simulation. This requires adding the ability to generate utterances with stylistic variations to capture different user personalities while maintaining consistency in style and vocabulary over a single dialogue.

\bibliographystyle{acl_natbib}
\bibliography{emnlp2020}

\begin{thebibliography}{38}
\expandafter\ifx\csname natexlab\endcsname\relax\def\natexlab#1{#1}\fi

\bibitem[{Bapna et~al.(2017)Bapna, T{\"u}r, Hakkani-T{\"u}r, and
  Heck}]{bapna2017towards}
Ankur Bapna, Gokhan T{\"u}r, Dilek Hakkani-T{\"u}r, and Larry Heck. 2017.
\newblock \href {https://doi.org/10.21437/Interspeech.2017-518} {{Towards
  Zero-Shot Frame Semantic Parsing for Domain Scaling}}.
\newblock \emph{Proc. Interspeech 2017}, pages 2476--2480.

\bibitem[{Barzilay and McKeown(2005)}]{barzilay2005sentence}
Regina Barzilay and Kathleen~R McKeown. 2005.
\newblock \href {https://doi.org/10.1162/089120105774321091} {{Sentence Fusion
  for Multidocument News Summarization}}.
\newblock \emph{Computational Linguistics}, 31(3):297--328.

\bibitem[{Budzianowski et~al.(2018)Budzianowski, Wen, Tseng, Casanueva, Ultes,
  Ramadan, and Gasic}]{budzianowski2018multiwoz}
Pawe{\l} Budzianowski, Tsung-Hsien Wen, Bo-Hsiang Tseng, I{\~n}igo Casanueva,
  Stefan Ultes, Osman Ramadan, and Milica Gasic. 2018.
\newblock \href {https://doi.org/10.18653/v1/D18-1547} {{MultiWOZ-A Large-Scale
  Multi-Domain Wizard-of-Oz Dataset for Task-Oriented Dialogue Modelling}}.
\newblock In \emph{{Proceedings of the 2018 Conference on Empirical Methods in
  Natural Language Processing}}, pages 5016--5026.

\bibitem[{Cao et~al.(2018)Cao, Li, Li, and Wei}]{cao2018retrieve}
Ziqiang Cao, Wenjie Li, Sujian Li, and Furu Wei. 2018.
\newblock \href {https://doi.org/10.18653/v1/P18-1015} {{Retrieve, Rerank and
  Rewrite: Soft Template Based Neural Summarization}}.
\newblock In \emph{{Proceedings of the 56th Annual Meeting of the Association
  for Computational Linguistics (Volume 1: Long Papers)}}, pages 152--161.

\bibitem[{Chen et~al.(2019)Chen, Chen, Qin, Yan, and
  Wang}]{chen2019semantically}
Wenhu Chen, Jianshu Chen, Pengda Qin, Xifeng Yan, and William~Yang Wang. 2019.
\newblock \href {https://doi.org/10.18653/v1/P19-1360} {Semantically
  conditioned dialog response generation via hierarchical disentangled
  self-attention}.
\newblock In \emph{Proceedings of the 57th Annual Meeting of the Association
  for Computational Linguistics}, pages 3696--3709, Florence, Italy.
  Association for Computational Linguistics.

\bibitem[{Chen et~al.(2020)Chen, Eavani, Chen, Liu, and
  Wang}]{chen-etal-2020-shot}
Zhiyu Chen, Harini Eavani, Wenhu Chen, Yinyin Liu, and William~Yang Wang. 2020.
\newblock \href {https://doi.org/10.18653/v1/2020.acl-main.18} {Few-shot {NLG}
  with pre-trained language model}.
\newblock In \emph{Proceedings of the 58th Annual Meeting of the Association
  for Computational Linguistics}, pages 183--190, Online. Association for
  Computational Linguistics.

\bibitem[{Devlin et~al.(2019)Devlin, Chang, Lee, and
  Toutanova}]{devlin2019bert}
Jacob Devlin, Ming-Wei Chang, Kenton Lee, and Kristina Toutanova. 2019.
\newblock \href {https://doi.org/10.18653/v1/N19-1423} {{BERT: Pre-training of
  Deep Bidirectional Transformers for Language Understanding}}.
\newblock In \emph{{Proceedings of the 2019 Conference of the North American
  Chapter of the Association for Computational Linguistics: Human Language
  Technologies, Volume 1 (Long and Short Papers)}}, pages 4171--4186.

\bibitem[{Doddington(2002)}]{doddington2002automatic}
George Doddington. 2002.
\newblock \href {https://dl.acm.org/doi/abs/10.5555/1289189.1289273}
  {{Automatic evaluation of machine translation quality using n-gram
  co-occurrence statistics}}.
\newblock In \emph{{Proceedings of the second international conference on Human
  Language Technology Research}}, pages 138--145. Morgan Kaufmann Publishers
  Inc.

\bibitem[{Du{ {s}}ek et~al.(2019)Du{ {s}}ek, Howcroft, and
  Rieser}]{duvsek2019semantic}
Ond{ {r}}ej Du{ {s}}ek, David~M Howcroft, and Verena Rieser. 2019.
\newblock \href {https://doi.org/10.18653/v1/W19-8652} {{Semantic Noise Matters
  for Neural Natural Language Generation}}.
\newblock In \emph{{Proceedings of the 12th International Conference on Natural
  Language Generation}}, pages 421--426.

\bibitem[{Du{ {s}}ek and Jurcicek(2016{\natexlab{a}})}]{duvsek2016context}
Ond{ {r}}ej Du{ {s}}ek and Filip Jurcicek. 2016{\natexlab{a}}.
\newblock \href {https://doi.org/10.18653/v1/W16-3622} {{A Context-aware
  Natural Language Generator for Dialogue Systems}}.
\newblock In \emph{{Proceedings of the 17th Annual Meeting of the Special
  Interest Group on Discourse and Dialogue}}, pages 185--190.

\bibitem[{Du{ {s}}ek and Jurcicek(2016{\natexlab{b}})}]{duvsek2016sequence}
Ond{ {r}}ej Du{ {s}}ek and Filip Jurcicek. 2016{\natexlab{b}}.
\newblock \href {https://doi.org/10.18653/v1/P16-2008} {{Sequence-to-Sequence
  Generation for Spoken Dialogue via Deep Syntax Trees and Strings}}.
\newblock In \emph{{Proceedings of the 54th Annual Meeting of the Association
  for Computational Linguistics (Volume 2: Short Papers)}}, pages 45--51.

\bibitem[{Du{\v{s}}ek and Jurcicek(2019)}]{duvsekneural}
Ond{\v{r}}ej Du{\v{s}}ek and Filip Jurcicek. 2019.
\newblock \href {https://doi.org/10.18653/v1/W19-8670} {{Neural Generation for
  Czech: Data and Baselines}}.
\newblock In \emph{Proceedings of the 12th International Conference on Natural
  Language Generation}, pages 563--574.

\bibitem[{Gardent et~al.(2017)Gardent, Shimorina, Narayan, and
  Perez-Beltrachini}]{gardent2017webnlg}
Claire Gardent, Anastasia Shimorina, Shashi Narayan, and Laura
  Perez-Beltrachini. 2017.
\newblock \href {https://doi.org/10.18653/v1/W17-3518} {{The WebNLG Challenge:
  Generating Text from RDF Data}}.
\newblock In \emph{{Proceedings of the 10th International Conference on Natural
  Language Generation}}, pages 124--133.

\bibitem[{Guu et~al.(2018)Guu, Hashimoto, Oren, and Liang}]{guu2018generating}
Kelvin Guu, Tatsunori~B Hashimoto, Yonatan Oren, and Percy Liang. 2018.
\newblock \href {https://doi.org/10.1162/tacl_a_00030} {{Generating Sentences
  by Editing Prototypes}}.
\newblock \emph{Transactions of the Association for Computational Linguistics},
  6:437--450.

\bibitem[{Hossain et~al.(2020)Hossain, Ghazvininejad, and
  Zettlemoyer}]{hossain2020simple}
Nabil Hossain, Marjan Ghazvininejad, and Luke Zettlemoyer. 2020.
\newblock \href {https://doi.org/10.18653/v1/2020.acl-main.228} {{Simple and
  Effective Retrieve-Edit-Rerank Text Generation}}.
\newblock In \emph{{Proceedings of the 58th Annual Meeting of the Association
  for Computational Linguistics}}, pages 2532--2538.

\bibitem[{Kale and Rastogi(2020)}]{kale2020text}
Mihir Kale and Abhinav Rastogi. 2020.
\newblock \href {https://arxiv.org/abs/2005.10433} {{Text-to-Text Pre-Training
  for Data-to-Text Tasks}}.
\newblock \emph{arXiv}, pages arXiv--2005.

\bibitem[{Kale and Roy(2020)}]{kale2020machine}
Mihir Kale and Scott Roy. 2020.
\newblock \href {https://arxiv.org/abs/2004.02077} {{Machine Translation
  Pre-training for Data-to-Text Generation--A Case Study in Czech}}.
\newblock \emph{arXiv preprint arXiv:2004.02077}.

\bibitem[{Keskar et~al.(2019)Keskar, McCann, Varshney, Xiong, and
  Socher}]{keskar2019ctrl}
Nitish~Shirish Keskar, Bryan McCann, Lav~R Varshney, Caiming Xiong, and Richard
  Socher. 2019.
\newblock \href {http://arxiv.org/abs/1909.05858} {{CTRL: A Conditional
  Transformer Language Model for Controllable Generation}}.
\newblock \emph{arXiv preprint arXiv:1909.05858}.

\bibitem[{Lavie and Agarwal(2007)}]{lavie2007meteor}
Alon Lavie and Abhaya Agarwal. 2007.
\newblock \href {https://www.aclweb.org/anthology/W07-0734} {{METEOR: An
  Automatic Metric for MT Evaluation with High Levels of Correlation with Human
  Judgments}}.
\newblock In \emph{{Proceedings of the Second Workshop on Statistical Machine
  Translation}}, pages 228--231. Association for Computational Linguistics.

\bibitem[{Lebret et~al.(2016)Lebret, Grangier, and Auli}]{lebret2016neural}
R{'e}mi Lebret, David Grangier, and Michael Auli. 2016.
\newblock \href {https://doi.org/10.18653/v1/D16-1128} {{Neural Text Generation
  from Structured Data with Application to the Biography Domain}}.
\newblock In \emph{{Proceedings of the 2016 Conference on Empirical Methods in
  Natural Language Processing}}, pages 1203--1213.

\bibitem[{Lin(2004)}]{lin2004rouge}
Chin-Yew Lin. 2004.
\newblock \href {https://www.aclweb.org/anthology/W04-1013} {{ROUGE: A Package
  for Automatic Evaluation of Summaries}}.
\newblock In \emph{{Text summarization branches out}}, pages 74--81.

\bibitem[{Liu et~al.(2019)Liu, Ott, Goyal, Du, Joshi, Chen, Levy, Lewis,
  Zettlemoyer, and Stoyanov}]{liu2019roberta}
Yinhan Liu, Myle Ott, Naman Goyal, Jingfei Du, Mandar Joshi, Danqi Chen, Omer
  Levy, Mike Lewis, Luke Zettlemoyer, and Veselin Stoyanov. 2019.
\newblock \href {http://arxiv.org/abs/1907.11692} {{RoBERTa: A Robustly
  Optimized BERT Pretraining Approach}}.
\newblock \emph{arXiv preprint arXiv:1907.11692}.

\bibitem[{Mi et~al.(2019)Mi, Huang, Zhang, and Faltings}]{mi2019meta}
Fei Mi, Minlie Huang, Jiyong Zhang, and Boi Faltings. 2019.
\newblock \href {https://doi.org/10.24963/ijcai.2019/437} {{Meta-Learning for
  Low-resource Natural Language Generation in Task-oriented Dialogue Systems}}.
\newblock In \emph{{Proceedings of the 28th International Joint Conference on
  Artificial Intelligence}}, pages 3151--3157. AAAI Press.

\bibitem[{Nayak et~al.(2017)Nayak, Hakkani-T{\"u}r, Walker, and
  Heck}]{nayak2017plan}
Neha Nayak, Dilek Hakkani-T{\"u}r, Marilyn Walker, and Larry Heck. 2017.
\newblock \href {https://doi.org/10.21437/Interspeech.2017-1525} {{To Plan or
  not to Plan? Discourse Planning in Slot-Value Informed Sequence to Sequence
  Models for Language Generation}}.
\newblock \emph{Proc. Interspeech 2017}, pages 3339--3343.

\bibitem[{Novikova et~al.(2017)Novikova, Du{ {s}}ek, and
  Rieser}]{novikova2017e2e}
Jekaterina Novikova, Ond{ {r}}ej Du{ {s}}ek, and Verena Rieser. 2017.
\newblock \href {https://doi.org/10.18653/v1/W17-5525} {{The E2E Dataset: New
  Challenges For End-to-End Generation}}.
\newblock In \emph{{Proceedings of the 18th Annual SIGdial Meeting on Discourse
  and Dialogue}}, pages 201--206.

\bibitem[{Papineni et~al.(2002)Papineni, Roukos, Ward, and
  Zhu}]{papineni2002bleu}
Kishore Papineni, Salim Roukos, Todd Ward, and Wei-Jing Zhu. 2002.
\newblock \href {https://doi.org/10.3115/1073083.1073135} {{BLEU: a Method for
  Automatic Evaluation of Machine Translation}}.
\newblock In \emph{{Proceedings of the 40th annual meeting on association for
  computational linguistics}}, pages 311--318. Association for Computational
  Linguistics.

\bibitem[{Peng et~al.(2020)Peng, Zhu, Li, Li, Li, Zeng, and Gao}]{peng2020few}
Baolin Peng, Chenguang Zhu, Chunyuan Li, Xiujun Li, Jinchao Li, Michael Zeng,
  and Jianfeng Gao. 2020.
\newblock \href {https://arxiv.org/abs/2002.12328} {{Few-shot Natural Language
  Generation for Task-Oriented Dialog}}.
\newblock \emph{arXiv preprint arXiv:2002.12328}.

\bibitem[{Radford et~al.(2019)Radford, Wu, Child, Luan, Amodei, and
  Sutskever}]{radford2019language}
Alec Radford, Jeffrey Wu, Rewon Child, David Luan, Dario Amodei, and Ilya
  Sutskever. 2019.
\newblock \href
  {https://d4mucfpksywv.cloudfront.net/better-language-models/language-models.pdf}
  {{Language Models are Unsupervised Multitask Learners}}.
\newblock \emph{OpenAI Blog}, 1(8):9.

\bibitem[{Raffel et~al.(2020)Raffel, Shazeer, Roberts, Lee, Narang, Matena,
  Zhou, Li, and Liu}]{raffel2020exploring}
Colin Raffel, Noam Shazeer, Adam Roberts, Katherine Lee, Sharan Narang, Michael
  Matena, Yanqi Zhou, Wei Li, and Peter~J Liu. 2020.
\newblock \href {http://jmlr.org/papers/v21/20-074.html} {{Exploring the Limits
  of Transfer Learning with a Unified Text-to-Text Transformer}}.
\newblock \emph{Journal of Machine Learning Research}, 21(140):1--67.

\bibitem[{Rastogi et~al.(2019)Rastogi, Zang, Sunkara, Gupta, and
  Khaitan}]{rastogi2019scalable}
Abhinav Rastogi, Xiaoxue Zang, Srinivas Sunkara, Raghav Gupta, and Pranav
  Khaitan. 2019.
\newblock \href {https://doi.org/10.1609/aaai.v34i05.6394} {{Towards Scalable
  Multi-domain Conversational Agents: The Schema-Guided Dialogue Dataset}}.
\newblock In \emph{{Proceedings of the AAAI Conference on Artificial
  Intelligence}}.

\bibitem[{Tran and Le~Nguyen(2018)}]{tran2018adversarial}
Van-Khanh Tran and Minh Le~Nguyen. 2018.
\newblock \href {https://www.aclweb.org/anthology/C18-1103} {{Adversarial
  Domain Adaptation for Variational Neural Language Generation in Dialogue
  Systems}}.
\newblock In \emph{{Proceedings of the 27th International Conference on
  Computational Linguistics}}, pages 1205--1217.

\bibitem[{Vedantam et~al.(2015)Vedantam, Lawrence~Zitnick, and
  Parikh}]{vedantam2015cider}
Ramakrishna Vedantam, C~Lawrence~Zitnick, and Devi Parikh. 2015.
\newblock \href
  {https://www.cv-foundation.org/openaccess/content_cvpr_2015/html/Vedantam_CIDEr_Consensus-Based_Image_2015_CVPR_paper.html}
  {{CIDEr: Consensus-based Image Description Evaluation}}.
\newblock In \emph{{Proceedings of the IEEE conference on computer vision and
  pattern recognition}}, pages 4566--4575.

\bibitem[{Wen et~al.(2016)Wen, Gasic, Mrk{ {s}}i{'c}, Barahona, Su, Vandyke,
  and Young}]{wen2016multi}
Tsung-Hsien Wen, Milica Gasic, Nikola Mrk{ {s}}i{'c}, Lina M~Rojas Barahona,
  Pei-Hao Su, David Vandyke, and Steve Young. 2016.
\newblock \href {https://doi.org/10.18653/v1/N16-1015} {{Multi-domain Neural
  Network Language Generation for Spoken Dialogue Systems}}.
\newblock In \emph{{Proceedings of the 2016 Conference of the North American
  Chapter of the Association for Computational Linguistics: Human Language
  Technologies}}, pages 120--129.

\bibitem[{Wen et~al.(2015)Wen, Gasic, Mrk{ {s}}i{'c}, Su, Vandyke, and
  Young}]{wen2015semantically}
Tsung-Hsien Wen, Milica Gasic, Nikola Mrk{ {s}}i{'c}, Pei-Hao Su, David
  Vandyke, and Steve Young. 2015.
\newblock \href {https://doi.org/10.18653/v1/D15-1199} {{Semantically
  Conditioned LSTM-based Natural Language Generation for Spoken Dialogue
  Systems}}.
\newblock In \emph{{Proceedings of the 2015 Conference on Empirical Methods in
  Natural Language Processing}}, pages 1711--1721.

\bibitem[{Wen et~al.(2017)Wen, Vandyke, Mrk{ {s}}i{'c}, Gasic, Barahona, Su,
  Ultes, and Young}]{wen2017network}
Tsung-Hsien Wen, David Vandyke, Nikola Mrk{ {s}}i{'c}, Milica Gasic, Lina
  M~Rojas Barahona, Pei-Hao Su, Stefan Ultes, and Steve Young. 2017.
\newblock \href {https://www.aclweb.org/anthology/E17-1042} {{A Network-based
  End-to-End Trainable Task-oriented Dialogue System}}.
\newblock In \emph{{Proceedings of the 15th Conference of the European Chapter
  of the Association for Computational Linguistics: Volume 1, Long Papers}},
  pages 438--449.

\bibitem[{Wu et~al.(2019)Wu, Wei, Huang, Wang, Li, and Zhou}]{wu2019response}
Yu~Wu, Furu Wei, Shaohan Huang, Yunli Wang, Zhoujun Li, and Ming Zhou. 2019.
\newblock \href {https://doi.org/10.1609/aaai.v33i01.33017281} {{Response
  Generation by Context-Aware Prototype Editing}}.
\newblock In \emph{{Proceedings of the AAAI Conference on Artificial
  Intelligence}}, volume~33, pages 7281--7288.

\bibitem[{Yang et~al.(2019)Yang, Dai, Yang, Carbonell, Salakhutdinov, and
  Le}]{yang2019xlnet}
Zhilin Yang, Zihang Dai, Yiming Yang, Jaime Carbonell, Russ~R Salakhutdinov,
  and Quoc~V Le. 2019.
\newblock \href
  {https://papers.nips.cc/paper/8812-xlnet-generalized-autoregressive-pretraining-for-language-understanding}
  {{XLNet: Generalized Autoregressive Pretraining for Language Understanding}}.
\newblock In \emph{{Advances in neural information processing systems}}, pages
  5754--5764.

\bibitem[{Zhu et~al.(2019)Zhu, Zeng, and Huang}]{zhu2019multi}
Chenguang Zhu, Michael Zeng, and Xuedong Huang. 2019.
\newblock \href {https://doi.org/10.18653/v1/D19-1123} {{Multi-task Learning
  for Natural Language Generation in Task-Oriented Dialogue}}.
\newblock In \emph{{Proceedings of the 2019 Conference on Empirical Methods in
  Natural Language Processing and the 9th International Joint Conference on
  Natural Language Processing (EMNLP-IJCNLP)}}, pages 1261--1266.

\end{thebibliography}

\maketitle

\appendix

\title{Appendix}

\date{}

\section{Additional Experiment Details}
\label{exp-details}
All models are trained on a 4x4 TPU slice, each taking ~1-3 hours to finish training for 5000 steps. We provide development set BLEU scores in Tables \ref{results-dev-sgd} and \ref{results-dev-fewshot}. These scores are computed on the entire development set which includes both seen and unseen domains. In Table \ref{results-test-fewshot}, we list the exact performance numbers for the few-shot NLG experiments.

\section{Automatic Metrics}
\label{appendix-metrics}
Prior work has used different metrics for different benchmarks. Moreover, for the same metric (e.g. BLEU), different implementations are used. For fair comparison, for each dataset, we report the results using the implementation used in prior work. For E2E, we use the implementation from the e2e-metrics \footnote{https://github.com/tuetschek/e2e-metrics} suite. For computing BLEU on MultiWOZ, we use code made available in the SC-GPT codebase \footnote{https://github.com/pengbaolin/SC-GPT}.
For model development i.e checking the best checkpoint based on the validation set, we rely on \textsl{sacrebleu} \footnote{https://github.com/mjpost/sacreBLEU} across all experiments, since it has become the standard implementation in machine translation literature. We urge the NLG community to also converge upon a single implementation of BLEU. Taking inspiration from MT, the BLEU scores on experiments involving the SGD dataset are computed using \textsl{sacrebleu}.

\begin{table}[]
\centering
\begin{tabular}{ll} \hline
model & BLEU           \\ \hline
Naive & 28.8 \\
SG    & 29.9 \\
T2G2  & 30.3 \\ \hline  
\end{tabular}
\caption{Development set performance on the SGD dataset.}
\label{results-dev-sgd}
\end{table}

\begin{table}[]
\centering
\begin{tabular}{cccc} \hline
\multicolumn{1}{l}{K} & \multicolumn{1}{l}{Naive} & \multicolumn{1}{l}{Schema} & \multicolumn{1}{l}{T2G2} \\ \hline
5                & 19.8                      & 20.0                       & 22.0                     \\
10               & 21.3                      & 22.0                       & 24.0                     \\
20               & 23.4                      & 22.4                       & 24.5                     \\
40               & 23.1                      & 25.3                       & 25.6                     \\
80               & 26.1                      & 24.9                       & 27.8                     \\
All              & 28.8                      & 29.9                       & 27.5      \\ \hline               
\end{tabular}
\caption{Development set BLEU scores in few-shot settings.  $K$-shot denotes $K$ dialogs for an API in the training set.}
\label{results-dev-fewshot}
\end{table}

\begin{table}[]
\centering
\begin{tabular}{ccccccc} \hline
\multicolumn{1}{l}{K} & \multicolumn{2}{c}{Naive}                          & \multicolumn{2}{c}{Schema}                         & \multicolumn{2}{c}{T2G2}                           \\ \hline
\multicolumn{1}{l}{}       & \multicolumn{1}{l}{BLEU} & \multicolumn{1}{l}{SER} & \multicolumn{1}{l}{BLEU} & \multicolumn{1}{l}{SER} & \multicolumn{1}{l}{BLEU} & \multicolumn{1}{l}{SER} \\ \hline
5                          & 18.7                     & 6.4                     & 18.9                     & 7.4                     & 23.8                     & 3.6                     \\
10                         & 19.7                     & 4.7                     & 19.5                     & 5.6                     & 24.4                     & 2.9                     \\
20                         & 20.6                     & 3.6                     & 20.4                     & 4.7                     & 24.7                     & 2.8                     \\
40                         & 21.4                     & 2.9                     & 21.4                     & 3.0                     & 26.0                     & 1.4                     \\
80                         & 23.0                     & 2.2                     & 21.7                     & 2.7                     & 27.8                     & 0.5                     \\
All                        & 26.3                     & 1.0                     & 26.2                     & 0.8                     & 28.6                     & 0.4 \\ \hline                   
\end{tabular}
\caption{Test set performance in few-shot settings. $K$-shot denotes $K$ dialogs for an API in the training set.}
\label{results-test-fewshot}
\end{table}

\section{Transfer Learning Across Domains}
\label{transfer-learning}
To measure the amount of transfer learning from one domain to another, we evaluate each domain specific model trained in Section \ref{joint-modeling} on all the domains and observe domain specific metrics.
Results can be found in Table \ref{results-domains-ser} and \ref{results-domains-bleu}.
\begin{table*}
\small
\centering
\begin{tabular}{l|l|l|l|l|l|l|l|l|l|l|l|l}
\hline
         & homes & buses & media & rides & movies & flights & music & services & rental & restaurants & events & hotels \\ \hline
homes       & 1.6   & 14.7  & 7.7   & 6.2         & 11.7   & 17.1    & 28.3  & 20       & 17.1       & 18          & 27.9   & 12.6   \\ \hline
buses       & 13.8  & 4     & 19.8  & 2.9         & 26.4   & 19.2    & 32.4  & 24.5     & 22.9       & 21.2        & 30.1   & 19     \\ \hline
media       & 38.6  & 42.4  & 8.4   & 20.2        & 33.6   & 48.7    & 26.9  & 44.6     & 38.7       & 36.7        & 40.8   & 37.4   \\ \hline
rides & 34.9  & 31.8  & 19.4  & 2.1         & 37.3   & 43.9    & 41.4  & 37.6     & 34.1       & 28.8        & 35.5   & 31.1   \\ \hline
movies      & 24.5  & 32.8  & 11    & 7.1         & 23     & 35.7    & 22.8  & 24.8     & 24.4       & 22.6        & 30.2   & 20.2   \\ \hline
flights     & 9.7   & 5.1   & 17    & 2.2         & 22.3   & 1       & 25.9  & 18.1     & 8.2        & 14.4        & 21.3   & 19.1   \\ \hline
music       & 36.6  & 38.5  & 3.9   & 20.1        & 24     & 48.4    & 0.6   & 26.3     & 28.4       & 23.9        & 38.3   & 33.6   \\ \hline
services    & 4.8   & 19.1  & 3.7   & 5.8         & 10.4   & 29.6    & 20.8  & 0.6      & 20.6       & 5.8         & 16.4   & 11.6   \\ \hline
rental  & 17.8  & 7     & 15.5  & 5.6         & 21.2   & 15.4    & 28.7  & 19.7     & 9.2        & 16.6        & 22.8   & 19.7   \\ \hline
restaurants & 9.8   & 21.9  & 10.9  & 5.2         & 21.9   & 33.1    & 24.7  & 6.9      & 18.4       & 4           & 15.7   & 19.2   \\ \hline
events      & 1.4   & 30.4  & 3.7   & 1.3         & 10     & 32.2    & 14.4  & 8.3      & 20.7       & 10          & 0.5    & 13.4   \\ \hline
hotels      & 5.2   & 10.1  & 6.3   & 1.3         & 8.8    & 19.8    & 18.6  & 5.2      & 6.7        & 6.3         & 8.5    & 1.6    \\ \hline
\end{tabular}
\caption{SER scores for domain specific models, when evaluated on all domains. The column denotes the domain on which the model was trained, while the row represents the domain used for evaluation.}
\label{results-domains-ser}
\end{table*}

\begin{table*}[]
\small
\begin{tabular}{l|l|l|l|l|l|l|l|l|l|l|l|l}
\hline
        & homes & buses & media & ridesg & movies & flights & music & services & rental & restaurants & events & hotels \\ \hline
homes       & 22.9  & 7.4   & 17.5  & 11.6        & 18.4   & 7.6     & 6.3   & 15.8     & 10.5       & 12.7        & 17     & 15.8   \\ \hline
buses       & 12.6  & 18.6  & 11.2  & 11.2        & 11.3   & 9.7     & 4.6   & 13       & 12         & 12          & 17.5   & 12.9   \\ \hline
media       & 6.1   & 5.6   & 28.9  & 9.1         & 16.2   & 3.8     & 10.6  & 9.5      & 4.9        & 8.7         & 8.4    & 11.5   \\ \hline
rides & 6.8   & 4.7   & 11.6  & 20.3        & 9.2    & 3.1     & 5.1   & 7.6      & 6.1        & 8.5         & 7.6    & 12.3   \\ \hline
movies      & 9.6   & 7.5   & 21    & 9.3         & 21.4   & 7.3     & 9.9   & 14       & 9.5        & 11.5        & 15.1   & 15.7   \\ \hline
flights     & 11.5  & 13.1  & 12.6  & 10.7        & 13.5   & 19.7    & 6     & 13       & 12.9       & 11.2        & 16.1   & 11.8   \\ \hline
music       & 8.5   & 5.3   & 21.7  & 8.3         & 17.9   & 3.9     & 25.3  & 11.2     & 5.2        & 9.6         & 10.9   & 12.1   \\ \hline
services    & 14.8  & 10.7  & 18.7  & 9.9         & 21     & 7.5     & 9.5   & 25.3     & 13.7       & 20.5        & 20.9   & 18.8   \\ \hline
rental  & 11.7  & 11.9  & 12    & 9.6         & 14.1   & 7.9     & 4.5   & 15       & 17.6       & 14.2        & 16.8   & 14.3   \\ \hline
restaurants & 15.4  & 10    & 17.4  & 10.5        & 17.1   & 8       & 9.5   & 21.2     & 12         & 25.8        & 19.3   & 17.9   \\ \hline
events      & 17.4  & 11.4  & 19.3  & 12.6        & 23.5   & 10.2    & 10.9  & 19.8     & 14         & 19.4        & 30.7   & 19.1   \\ \hline
hotels      & 12.1  & 9.1   & 15.6  & 8.7         & 18.9   & 8       & 6.9   & 17.2     & 10.5       & 16.8        & 17.3   & 26.6   \\ \hline
\end{tabular}
\caption{BLEU scores for domain specific models, when evaluated on all domains. The column denotes the domain on which the model was trained, while the row represents the domain used for evaluation.}
\label{results-domains-bleu}
\end{table*}

\section{Templates}
\label{template-examples}
In Tables \ref{templates-travel}, \ref{templates-ridesharing} and \ref{templates-movies}, we provide templates used for a few different APIs. The full set of templates is available with the code. Note that the linguistic quality of the templates does not need to be very high, as long as the semantics of the dialog act are captured. This makes it easy for the API developers themselves to quickly create the simple templates.

\section{Human Evaluation Tasks}
\label{human-eval-tasks}
\begin{figure*}[t]
    \centering
    \includegraphics[width=1.0\textwidth]{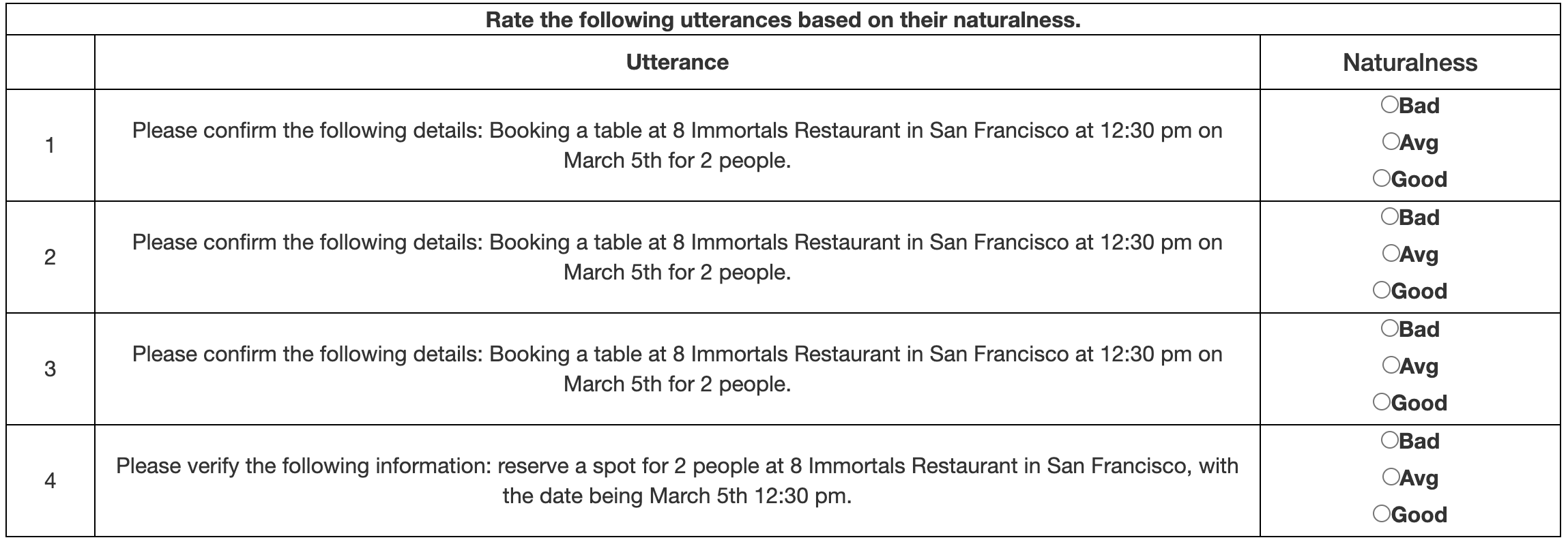}
    \caption{Example of a human rater task to evaluate naturalness. Each row represents the output from one of Naive, Schema, T2G2 and Ground Truth. The order of rows is shuffled across different tasks.}
    \label{fig:eval_nat}
\end{figure*}

\begin{figure*}
    \centering
    \includegraphics[width=1.0\textwidth]{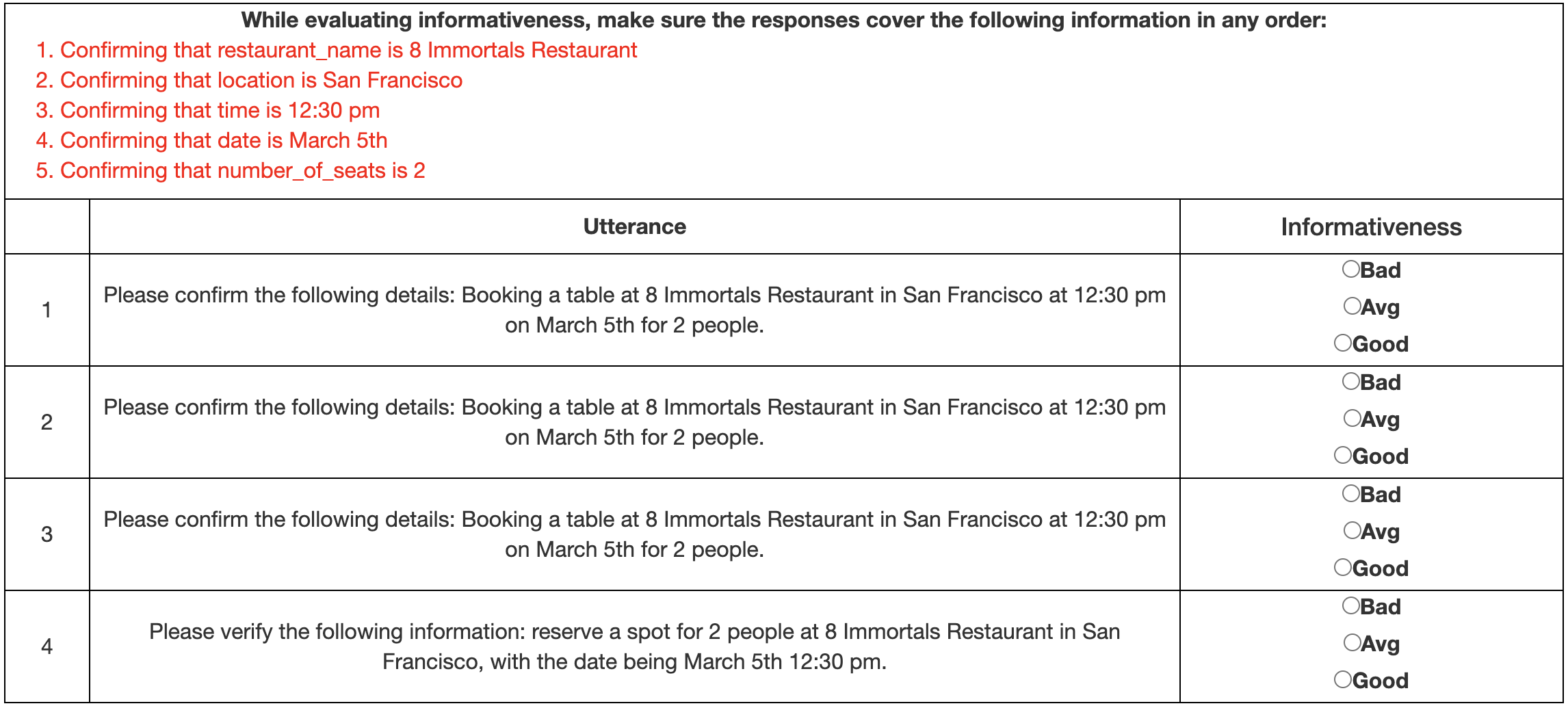}
    \caption{Example of a human rater task to evaluate informativeness. Each row represents the output from one of Naive, Schema, T2G2 and Ground Truth. The order of rows is shuffled across different tasks.}
    \label{fig:eval_info}
\end{figure*}

Figures \ref{fig:eval_nat} and \ref{fig:eval_info} show examples of rater tasks for naturalness and informativeness respectively.

\begin{table*}[]
\begin{tabular}{|l|l|} \hline
\textbf{Dialog Act} & \textbf{Template} \\ \hline
GOODBYE                                  & Enjoy!                                              \\
INFORM!!free\_entry!!False               & No, entry to this place is not free.                \\
INFORM!!free\_entry!!True                & Yes, entry is free.                                 \\
INFORM!!good\_for\_kids!!False           & No, this place isn't exactly good for kids.         \\
INFORM!!good\_for\_kids!!True            & Yes, the place is pretty nice to take your kids to. \\
INFORM!!phone\_number!!@                 & Their phone number is @.                            \\
INFORM\_COUNT!!FindAttractions!!count!!@ & There are @ attractions that could interest you.    \\
NOTIFY\_FAILURE!!FindAttractions         & Sorry, can't find anything for your constraints.    \\
OFFER!!attraction\_name!!@               & You should check out @.                             \\
OFFER!!category!!@                       & This is a @.                                        \\
REQUEST!!location                        & Where should I search for attractions in?           \\
REQ\_MORE                                & Anything else I could assist with?          \\ \hline       
\end{tabular}
\caption{Templates from a Travel API.}
\label{templates-travel}
\end{table*}

\begin{table*}[]
\begin{tabular}{|l|l|} \hline
\textbf{Dialog Ac}t & \textbf{Template} \\ \hline
CONFIRM!!destination!!@       & You want a ride to @.                                                      \\
CONFIRM!!number\_of\_seats!!@ & You need @ seats.                                                          \\
CONFIRM!!ride\_type!!@        & You prefer a @ ride.                                                       \\
GOODBYE                       & Have a wonderful day.                                                      \\
INFORM!!ride\_fare!!@         & The ride costs @.                                                          \\
INFORM!!wait\_time!!@         & The cab is expected in @ minutes.                                          \\
NOTIFY\_FAILURE!!GetRide      & I'm sorry, I could not find a ride for you at this time.                   \\
NOTIFY\_SUCCESS!!GetRide      & I booked your ride and the cab is on its way.                              \\
REQUEST!!destination          & Where do you want to go to?                                                \\
REQUEST!!destination!!@       & Are you going to @?                                                        \\
REQUEST!!destination!!@@      & Are you going to @ or @?                                                   \\
REQUEST!!number\_of\_seats    & How many seats do you need?                                                \\
REQUEST!!ride\_type           & Do you have a preferred type of ride?                                      \\
REQ\_MORE                     & Can I help you with anything else? \\ \hline                            
\end{tabular}
\caption{Templates from a RideSharing API.}
\label{templates-ridesharing}
\end{table*}

\begin{table*}[]
\begin{tabular}{|l|l|} \hline
\textbf{Dialog Ac}t & \textbf{Template} \\ \hline
CONFIRM!!subtitle\_language!!@      & with subtitles in @.                                  \\
CONFIRM!!title!!@                   & playing @                                             \\
GOODBYE                             & Have a good day.                                      \\
INFORM!!genre!!@                    & It is a @ movie.                                      \\
INFORM!!starring!!@                 & @ acted in it.                                        \\
INFORM\_COUNT!!FindMovies!!count!!@ & There're @ movies you may like.                       \\
NOTIFY\_FAILURE!!FindMovies         & I failed to find any movies matching your preference. \\
NOTIFY\_FAILURE!!PlayMovie          & Failed to play the movie.                             \\
NOTIFY\_SUCCESS!!PlayMovie          & Started playing the movie.                            \\
OFFER!!title!!@                     & What about @?                                         \\
OFFER!!title!!@@                    & What about @ or @?                                    \\
OFFER!!title!!@@@                   & Do you like @, @ or @?                                \\
OFFER\_INTENT!!PlayMovie            & Do you want to play the movie?                        \\
REQUEST!!genre                      & What kind of movies do you like?                      \\
REQUEST!!title                      & Which movie do you want to watch?                     \\
REQ\_MORE     & What else can I help?  \\ \hline                            
\end{tabular}
\caption{Templates from a Movies API.}
\label{templates-movies}
\end{table*}

\section{Example Outputs}
\label{qualitative-examples}
Sample utterances generated using the different models for various domains are shown in the examples below. The system actions, its template based representation used by the T2G2 model as input, and the reference response are also provided. The predictions are from models trained on the full SGD dataset and without any dialogue history context. The unseen domains have been marked with an asterisk.

\begin{figure*}[h!]\centering
\begin{minipage}{15.6cm}\vspace{0mm}    \centering
\begin{tcolorbox}[colback=yellow!5!white]
    \centering
    \scriptsize
    \begin{tabular}{llp{0.70\columnwidth}}
    \textbf{Domain} & \textbf{Model} & \textbf{Input or Generated sequence} \\
    \midrule
    \textbf{RentalCars} & \textit{Actions} & request(start\_date=March 12th) request(pickup\_location) \\[0.5pt]
    & \textit{Template} & Do you want to have the car from March 12th? Where do you want to pick up the car? \\[0.5pt]
    & \textit{Reference} & Will you be renting the car from March 12th? Please provide the pick-up location as well. \\[2.5pt]
    & \textbf{Naive} & When do you want to start on March 12th and where do you want to pick it up? \\[0.5pt]
    & \textbf{SlotDesc} & When do you want it from March 12th to pick it up? \\[0.5pt]
    & \textbf{T2G2} & Do you want to pick up the car from March 12th and where do you want to pick it up? \\[2.5pt]
    \midrule
    \textbf{Hotels} & \textit{Actions} & inform(smoking\_allowed=False) notify\_success \\[0.5pt]
    & \textit{Template} & Successfully booked the hotel. No, smoking is not allowed on the property. \\[0.5pt]
    & \textit{Reference} & Your reservation is booked. No, smoking is not allowed on the property at this hotel. \\[2.5pt]
    & \textbf{Naive} & Your reservation has been made. Unfortunately, the hotel does not allow you to make the reservation. \\[0.5pt]
    & \textbf{SlotDesc} & Your reservation has been made. Unfortunately, they do not allow smoking. \\[0.5pt]
    & \textbf{T2G2} & I have booked the hotel. No, smoking is not allowed on the property. \\[2.5pt]
    \midrule
    \textbf{Homes} & \textit{Actions} & confirm(property\_name=Almaden lake Apartments) confirm(visit\_date=March 13th) \\[0.5pt]
    & \textit{Template} & Please confirm the following details: You are scheduling a visit to Almaden Lake Apartments and you want to visit the property on March 13th. \\[0.5pt]
    & \textit{Reference} & Please confirm details to schedule a visit to Almaden lake Apartments on March 13th. \\[2.5pt]
    & \textbf{Naive} & Please confirm the following details: You are scheduling a visit to Almaden lake Apartments on March 13th. \\[0.5pt]
    & \textbf{SlotDesc} & Please confirm: You want to visit Almaden lake Apartments on March 13th. \\[0.5pt]
    & \textbf{T2G2} & Please confirm the following details: You are scheduling a visit to Almaden Lake Apartments on March 13th. \\[2.5pt]
    \midrule
    \textbf{Services} & \textit{Actions} & request(type) \\[0.5pt]
    & \textit{Template} & What kind of therapist do you need? \\[0.5pt]
    & \textit{Reference} & Let me know your required type of Therapist to search for. \\[2.5pt]
    & \textbf{Naive} & What kind of doctor are you looking for? \\[0.5pt]
    & \textbf{SlotDesc} & What kind of therapist are you looking for? \\[0.5pt]
    & \textbf{T2G2} & What kind of therapist do you need? \\[2.5pt]
    \midrule
    \textbf{Restaurants} & \textit{Actions} & confirm(restaurant\_name=Nizza La Bella) confirm(location=Albany) confirm(time=6:15 pm) confirm(number\_of\_seats=2) confirm(date=March 7th) \\[0.5pt]
    & \textit{Template} & Please confirm the following details: Booking a table at Nizza La Bella. The city is Albany. The reservation is at 6:15 pm. The reservation is for 2 people. The date is March 7th. \\[0.5pt]
    & \textit{Reference} & Please confirm: You want me to book a table for 2 people at Nizza La Bella located in Albany for 6:15 pm on March 7th. \\[2.5pt]
    & \textbf{Naive} & Please confirm the following details: Booking a table at Nizza La Bella in Albany at 6:15 pm for 2 people on March 7th. \\[0.5pt]
    & \textbf{SlotDesc} & Please confirm the following details: Booking a table at Nizza La Bella in Albany at 6:15 pm for 2 people on March 7th. \\[0.5pt]
    & \textbf{T2G2} & Please confirm the following details: Booking a table at Nizza La Bella in Albany at 6:15 pm on March 7th for 2 people. \\[2.5pt]
    \midrule
    \textbf{Media} & \textit{Actions} & inform(starring=Antonio Bustorff) \\[0.5pt]
    & \textit{Template} & Antonio Bustorff acted in it. \\[0.5pt]
    & \textit{Reference} & The actor is Antonio Bustorff. \\[2.5pt]
    & \textbf{Naive} & The movie is directed by Antonio Bustorff. \\[0.5pt]
    & \textbf{SlotDesc} & The movie is directed by Antonio Bustorff. \\[0.5pt]
    & \textbf{T2G2} & Antonio Bustorff acted in the movie. \\[2.5pt]
    \midrule
    \textbf{Music} & \textit{Actions} & confirm(track=strangers) confirm(device=Patio) \\[0.5pt]
    & \textit{Template} & Please confirm the following details: playing the song Strangers in Patio \\[0.5pt]
    & \textit{Reference} & Please confirm: play strangers in patio? \\[2.5pt]
    & \textbf{Naive} & Please confirm the following details: Play strangers on the Patio. \\[0.5pt]
    & \textbf{SlotDesc} & Please confirm: play strangers on Patio. \\[0.5pt]
    & \textbf{T2G2} & Please confirm the following details: playing Strangers in Patio. \\[2.5pt]
    \midrule
    \textbf{Buses} & \textit{Actions} & request(from\_city) request(departure\_date) \\[0.5pt]
    & \textit{Template} & Where are you leaving from? When do you plan to leave? \\[0.5pt]
    & \textit{Reference} & What is your departure date and city? \\[2.5pt]
    & \textbf{Naive} & Where are you leaving from and when do you plan to leave? \\[0.5pt]
    & \textbf{SlotDesc} & Where are you leaving from and when do you plan to leave? \\[0.5pt]
    & \textbf{T2G2} & Where are you leaving from and when do you plan to leave? \\[2.5pt]
    
 \vspace{-5mm}
    \end{tabular}
\end{tcolorbox}
\vspace{-2mm}
\caption{A few examples of utterances generated from different models, along with the system actions, their template based representation (as shown in Figure~\ref{fig:overall}) and the reference output. The unseen domains have been marked with an asterisk.}
\label{fig:appendix-outputs1}
\end{minipage}
\end{figure*}

\begin{figure*}[h]\centering
\begin{minipage}{15.6cm}\vspace{0mm}    \centering
\begin{tcolorbox}[colback=yellow!5!white]
    \centering
    \scriptsize
    \begin{tabular}{llp{0.70\columnwidth}}
    \textbf{Domain} & \textbf{Model} & \textbf{Input or Generated sequence} \\
    \midrule
    \textbf{Events} & \textit{Actions} & confirm(event\_name=Acting For Teens) confirm(number\_of\_tickets=4) confirm(date=March 11th) confirm(city=New York) \\[0.5pt]
    & \textit{Template} & Please confirm the following details: Booking tickets for Acting For Teens. You have 4 tickets in your reservation. The date of it is March 11th. The event is taking place in New York. \\[0.5pt]
    & \textit{Reference} & Please confirm: You want me to purchase 4 tickets for the event called Acting For Teens taking place in New York on March 11th. \\[2.5pt]
    & \textbf{Naive} & Please confirm: 4 tickets to Acting For Teens on March 11th in New York. \\[0.5pt]
    & \textbf{SlotDesc} & Please confirm: 4 tickets to Acting For Teens on March 11th in New York. \\[0.5pt]
    & \textbf{T2G2} & Please confirm: Booking 4 tickets for Acting For Teens on March 11th in New York. \\[2.5pt]
    \midrule
    \textbf{Events} & \textit{Actions} & confirm(event\_name=Mxmtoon) confirm(number\_of\_tickets=2) confirm(date=March 10th) confirm(city=New York) \\[0.5pt]
    & \textit{Template} & Please confirm the following details: Booking tickets for Mxmtoon. You have 2 tickets in your reservation. The date of it is March 10th. The event is taking place in New York. \\[0.5pt]
    & \textit{Reference} & Can you please confirm that you want to buy 2 tickets for the Mxmtoon event which happens on March 10th in New York? \\[2.5pt]
    & \textbf{Naive} & Please confirm: 2 tickets for Mxmtoon on March 10th in New York. \\[0.5pt]
    & \textbf{SlotDesc} & Please confirm: 2 tickets to Mxmtoon on March 10th in New York. \\[0.5pt]
    & \textbf{T2G2} & Please confirm: 2 tickets for Mxmtoon on March 10th in New York. \\[2.5pt]
    \midrule
    \textbf{Alarm\textsuperscript{*}} & \textit{Actions} & offer(alarm\_time=4 pm) offer(alarm\_name=Event) \\[0.5pt]
    & \textit{Template} & There is an alarm for 4 pm with name Event. \\[0.5pt]
    & \textit{Reference} & A 4 pm alarm with name Event. \\[2.5pt]
    & \textbf{Naive} & There is an event at 4 pm. \\[0.5pt]
    & \textbf{SlotDesc} & There is an event at 4 pm. \\[0.5pt]
    & \textbf{T2G2} & There is an alarm for 4 pm with event name. \\[2.5pt]
    \midrule
    \textbf{Alarm\textsuperscript{*}} & \textit{Actions} & offer(alarm\_time=10:30 am) offer(alarm\_name=commute) inform\_count(count=1) \\[0.5pt]
    & \textit{Template} & You have 1 alarms currently set. There is an alarm for 10:30 am with name Commute. \\[0.5pt]
    & \textit{Reference} & You have 1 alarm at 10:30 am. with name commute \\[2.5pt]
    & \textbf{Naive} & I found 1 suitable for you. It's at 10:30 am and it's a commute. \\[0.5pt]
    & \textbf{SlotDesc} & I found 1 alarm at 10:30 am. It's a commute. \\[0.5pt]
    & \textbf{T2G2} & You have 1 alarm for 10:30 am with the name Commute. \\[2.5pt]
    \midrule
    \textbf{Weather} & \textit{Actions} & inform(humidity=30) \\[0.5pt]
    & \textit{Template} & The humidity is around 30 percent. \\[0.5pt]
    & \textit{Reference} & The humidity is about 30 percent. \\[2.5pt]
    & \textbf{Naive} & It's 30 minutes. \\[0.5pt]
    & \textbf{SlotDesc} & The humidity is 30. \\[0.5pt]
    & \textbf{T2G2} & The humidity is around 30 percent. \\[2.5pt]
    \midrule
    \textbf{Weather} & \textit{Actions} & inform(humidity=75) inform(wind=4) \\[0.5pt]
    & \textit{Template} & The humidity is around 75 percent. The average wind speed should be 4 miles per hour. \\[0.5pt]
    & \textit{Reference} & The humidity for today is around 75 percent. The average wind is 4 miles an hour. \\[2.5pt]
    & \textbf{Naive} & It's a 7 star, and it's a 4 star. \\[0.5pt]
    & \textbf{SlotDesc} & The humidity is 75 and the wind speed is 4 miles per hour. \\[0.5pt]
    & \textbf{T2G2} & The average wind speed should be 4 miles per hour and the humidity is around 75 percent. \\[2.5pt]
    \midrule
    \textbf{Messaging\textsuperscript{*}} & \textit{Actions} & notify\_success \\[0.5pt]
    & \textit{Template} & You've successfully shared the location. \\[0.5pt]
    & \textit{Reference} & The location was successfully shared. \\[2.5pt]
    & \textbf{Naive} & Your reservation has been made. \\[0.5pt]
    & \textbf{SlotDesc} & Your contact has been confirmed. \\[0.5pt]
    & \textbf{T2G2} & Your location has been shared. \\[2.5pt]
    \midrule
    \textbf{Messaging\textsuperscript{*}} & \textit{Actions} & confirm(location=2190 Bancroft Way) confirm(contact\_name=Peter) \\[0.5pt]
    & \textit{Template} & Please confirm the following details: You're sharing the location 2190 Bancroft Way with Peter. \\[0.5pt]
    & \textit{Reference} & So share the location 2190 Bancroft Way with Peter? \\[2.5pt]
    & \textbf{Naive} & Please confirm the following details: You are scheduling a visit to Peter at 2190 Bancroft Way. \\[0.5pt]
    & \textbf{SlotDesc} & Please confirm the following details: You would like to contact Peter at 2190 Bancroft Way. \\[0.5pt]
    & \textbf{T2G2} & Please confirm the following details: You're sharing the location 2190 Bancroft Way with Peter. \\[2.5pt]
    \midrule
    \textbf{Trains\textsuperscript{*}} & \textit{Actions} & offer(journey\_start\_time=7:10 am) offer(total=\$91) \\[0.5pt]
    & \textit{Template} & What about the train departing at 7:10 am? It costs \$91 in total. \\[0.5pt]
    & \textit{Reference} & There is a train leaving at 7:10 am and costs \$91. \\[2.5pt]
    & \textbf{Naive} & There is a bus that departs at 7:10 am and costs \$91. \\[0.5pt]
    & \textbf{SlotDesc} & There is a 7:10 am train that costs \$91. \\[0.5pt]
    & \textbf{T2G2} & How about the 7:10 am train? It costs \$91 in total. \\[2.5pt]
    \midrule
    \textbf{Travel} & \textit{Actions} & offer(attraction\_name=BODY WORLDS London) offer(category=Museum) \\[0.5pt]
    & \textit{Template} & You should check out BODY WORLDS London. This is a Museum. \\[0.5pt]
    & \textit{Reference} & I suggest a museum called BODY WORLDS London. \\[2.5pt]
    & \textbf{Naive} & BODY WORLDS London is a Museum. \\[0.5pt]
    & \textbf{SlotDesc} & BODY WORLDS London is a museum. \\[0.5pt]
    & \textbf{T2G2} & BODY WORLDS London is a museum. \\[2.5pt]
    
 \vspace{-5mm}
    \end{tabular}
\end{tcolorbox}
\vspace{-2mm}
\caption{Continuation of examples in Figure \ref{fig:appendix-outputs1}.}
\end{minipage}
\end{figure*}



\end{document}